\pdfoutput=1
\documentclass[11pt,a4paper]{article}
\usepackage[hyperref]{eacl2021}
\usepackage{times}
\usepackage{latexsym}

\usepackage{microtype}

\usepackage{amsmath}
\usepackage{amsfonts}
\usepackage{amssymb}
\usepackage{booktabs}
\usepackage{natbib}
\usepackage{graphicx}
\usepackage{enumitem}
\usepackage{mathtools}
\usepackage{framed}
\usepackage[utf8]{inputenc}
\usepackage{tikz}
\usetikzlibrary{shapes,arrows}
\usetikzlibrary{fit,positioning}
\usepackage{subcaption}
\usetikzlibrary{bayesnet}
\usepackage{arydshln}
\PassOptionsToPackage{linktocpage}{hyperref}

\aclfinalcopy 

\newcommand{\pink}[1]{%
  \begingroup
  \setlength{\fboxsep}{0pt}%
  \colorbox{pink!30}{\reducedstrut#1\/}%
  \endgroup
}
\newcommand{\green}[1]{%
  \begingroup
  \setlength{\fboxsep}{0pt}%
  \colorbox{green!30}{\reducedstrut#1\/}%
  \endgroup
}

\newcommand{\reducedstrut}{\vrule width 0pt height .9\ht\strutbox depth .9\dp\strutbox\relax}

\newcommand{\given}{\, \vert \,}

\title{Evaluating for Diversity in Question Generation over Text}

\author{Michael Sejr Schlichtkrull\thanks{\hspace{0.3cm}Work done as an intern at Amazon.}\\
  University of Amsterdam \\
  University of Edinburgh \\
  {\tt m.s.schlichtkrull@uva.nl}
  \And
  Weiwei Cheng\\
  Amazon \\
  {\tt weiweic@amazon.com}}

\date{}

\begin{document}
\maketitle
\begin{abstract}
Generating diverse and relevant questions over text is a task with widespread applications. We argue that commonly-used evaluation metrics such as BLEU and METEOR are not suitable for this task due to the inherent diversity of reference questions, and propose a scheme for extending conventional metrics to reflect diversity. We furthermore propose a variational encoder-decoder model for this task. We show through automatic and human evaluation that our variational model improves diversity without loss of quality, and demonstrate how our evaluation scheme reflects this improvement.
\end{abstract}

\section{Introduction}

Question generation has widespread applications in online education \citep{lindberg2013generating}, search \citep{rothe2016asking}, and question answering \citep{yang2017semi, lewis2018generative}. 
Generating a single question per context can be inadequate as questions are naturally diverse in aspect, answer, and phrasing. Following \citet{li2016diversity}, the first two correspond to \textit{semantic diversity}, while the third one corresponds to \textit{lexical diversity}.

Evaluation in question generation uses metrics including BLEU \citep{papineni2002bleu}, METEOR \citep{banerjee2005meteor}, and ROUGE \citep{lin2004rouge}. Common to these metrics is an implicit assumption that \textit{references are paraphrases}. In question generation this assumption does not hold, as multiple \textit{diverse} questions may be equally relevant for a given context. To measure the ability of a system to produce diverse sets of questions, a new evaluation metric is needed. We propose a function composition framework consisting of meta-metrics relying on existing evaluation metrics. As we will show in Section \ref{sec:measure}, our framework generalizes $F_1$, a widely-used measure in information retrieval and multivariate analysis.

\begin{figure}[htb!]
\begin{framed}\small
\textbf{Context:} Amazon.com, Inc. is an American electronic commerce and cloud computing company based in Seattle, Washington, founded by Jeff Bezos on July 5, 1994.

\medskip

\textbf{Question 1:} $\quad$ Which Seattle-based company was founded in 1994?

\hfill --- Amazon.com

\smallskip

\textbf{Question 2:} $\quad$ What is the name of the company that Jeff Bezos founded?

\hfill --- Amazon.com

\smallskip

\textbf{Question 3:} $\quad$ When was Amazon.com founded?

\hfill --- July 5, 1994



\end{framed}
\caption{Example context with questions naturally diverse in subject, structure, and answer.}
\label{figure:diversity_example}
\end{figure}

Recent papers on text generation over images have sought to increase diversity through generative modeling such as variational autoencoders \citep{jain2017creativity}. To demonstrate the usefulness of our F-like metrics, we build a conditional variational autoencoder for text-to-text generation inspired by these ideas and based on 
\citet{zhang2016variational}. We achieve significant improvements in diversity without performance loss, in terms of both automatic and human evaluation. 
We show that our F-like approach better captures the improvement in diversity offered by this model, and that the breakdown into ``precision'' and ``recall'' enabled by the meta-metric helps illustrate the strengths and weaknesses of different systems.


\section{Related Work}
Early work on natural language question generation focuses primarily on rule-based models \citep{heilman2010good, agarwal2011automatic} and template-based slot filling \citep{lindberg2013generating}. End-to-end neural models were introduced by \citet{du2017learning}, and extende in \citep{chan-fan-2019-bert, bao2020unilmv2}. Generating diverse questions is discussed in a series of recent papers \citep{zhou2017neural, harrison2018neural, song2018leveraging, yao2018teaching, shen2018medical} attempting to improve \textit{semantic diversity} by conditioning on (potential) answer positions or question types. While showing promising results, such prior information may not always be practically available. 
\citet{sultan-etal-2020-importance} recently applied nucleus sampling \citep{Holtzman2020The} to diversify question generation models, demonstrating improved performance on a downstream question answering task.

Diverse text generation has been studied for other tasks, including image question generation \citep{jain2017creativity}, conversation modeling \citep{li2016diversity}, machine translation \citep{zhang2016variational, schulz2018stochastic}, and image captioning \citep{vijayakumar2016diverse, pu2016variational, dai2017towards}. Models typically rely either on conditional variational autoencoders (CVAE) \citep{kingma2013auto, sohn2015learning} or conditional generative adversarial networks (CGAN) \citep{ mirza2014conditional}, and evaluation has relied on BLEU, ROUGE, and METEOR. 

Several schemes for evaluating with diversity in image captioning were proposed in \citep{alihosseini-etal-2019-jointly}. Their proposals rely either on statistical divergence between language modeling of the generated and reference sets, or on Jaccard index computed at the n-gram level. In \citet{dhingra-etal-2019-handling}, another metric is proposed for scoring systems where generated sentences overlapping with items on the \textit{source} side rather than the \textit{target} side should also be scored highly.



\section{Evaluating for Diversity}\label{sec:measure}

 
Conventional metrics evaluate the correctness of proposed questions in relation to a set of reference questions; they do not, however, measure the degree to which the proposed questions \textit{cover} the set of reference questions. Consider the sentence \textit{``Germany won the 2014 world cup"}, paired with two reference questions \textit{``Who won the 2014 world cup?"} and \textit{``Which event did Germany win in 2014?"}. Using traditional metrics, a system that always generates \textit{``Who won the 2014 world cup?"} and a system that alternates between generating the two would be given equal, perfect scores, when in fact the first system has only learned half the task. Moreover, systems that generate a sentence usings parts of both questions may wrongly be scored highly 
(see Appendix A for an example).


In this work, we propose a framework to extend commonly used scoring functions to account for coverage over reference questions. 
Given a set of reference questions $R \subset \Omega$, a set of predicted questions $P \subset \Omega$, and a scoring function $s: \Omega \times \Omega \to \mathbb{R}$, we propose the following two functions for comparing $P$ and $R$:
\begin{align}
u(P,R;s) &= \frac{1}{|P|} \sum\limits_{p \in P} \max\limits_{r \in R} s(p,r) \label{eq:u}\\
v(P,R;s) &= \frac{1}{|R|} \sum\limits_{r \in R} \max\limits_{p \in P} s(p,r) \label{eq:v}
\end{align}

To compute the function $u$, we identify for each predicted question the best match in the references w.r.t.\ the scoring function. We then compute how ``close'' the predictions are to the references by summing up scores between each predicted question and its respective best match. The function $v$ is computed analogously in Eq.\ (\ref{eq:v}).\footnote{Note that $v(P,R;s) \neq u(R,P;s)$ since $s$ is not necessarily symmetric, i.e., $s(x,y) \neq s(y,x)$.} We combine $u$ and $v$ with their harmonic mean. This leads to an overall measure to compare $P$ and $R$, 
\begin{align}\label{eq:f}
F(P,R) = \frac{2 \times u(P,R) \times v(P,R)}{u(P,R) + v(P,R)} \, .
\end{align}

In the special case of a binary scoring function $s: \Omega \times \Omega \to \{0,1\}$, where $1$ is given to exact matches,  $u$ and $v$ are identical to conventional \textit{precision} and \textit{recall}, respectively, and Eq.\ (\ref{eq:f}) is equal to $F_1$. Therefore, we can interpret $u$ and $v$ as generalized precision and recall, respectively.


To the best of our knowledge, our construction of the $F$ function is new. The ``best match'' idea used in comparing $P$ and $R$, i.e., examining the optimal score an item in the set can attain w.r.t.\ another set, has been applied in local community detection in network analysis \citep{clauset2005, lancichinetti2009} and behavior research in social networks \citep{adali2010}. In machine learning, the work by \citet{goldberg2010} on clustering analysis bears the closest resemblance to this idea. However, their assumption of a symmetric $s$-function does not hold for many NLP applications.

\section{Variational Question Generation}
Our approach extends the encoder-decoder model proposed by \citet{du2017learning} with a simple latent variable following \citet{zhang2016variational}. Given a training corpus of context-question pairs $\mathcal{D} = \{(X_1, Y_1), (X_2, Y_2), \ldots, (X_D, Y_D)\}$ such that $X_i = \{x_{i1}, \ldots, x_{ij}\}$ and $Y_i = \{y_{i1}, \ldots, y_{ik}\}$, the objective is to minimize the negative log-likelihood of the training data. We first encode each token in the context paragraph using either GloVe embeddings \citep{pennington2014glove} or ELMo embeddings \citep{peters2018elmo}. The entire sentence is then encoded through a 600-dimensional BiLSTM, giving forwards, backwards, and concatenated representations $\overrightarrow{b_t}$, $\overleftarrow{b_t}$, $b_t = \left[\overrightarrow{b_t}, \overleftarrow{b_t}\right]$ for each timestep $t$.

We construct context representations for each decoding step using bilinear attention as presented by \citet{luong2015effective}. That is, given a query vector $k_t$ corresponding to decoding step $t$:
\begin{align}
a_{i,t} &= \frac{\exp(k_t^T W_a b_i)}{\sum_j \exp(k_t^T W_a b_j)} \\
r_t &= \sum_i^{\lvert X \rvert} a_{i,t} b_i \\
c_t &= \tanh (W_c \left[r_t, k_t\right] )
\end{align}
We compute query vectors using another 
BiLSTM. The input to each step $t$ consists of the concatenation of the GloVe embedding $e_{t}$ of the target-side token generated at the previous timestep $t-1$, and the previous context vector $c_{t-1}$. That is
\begin{align}\label{eq:query}
k_t &= \text{LSTM}(\left[e_{t}, c_{t-1}\right], k_{t-1}) \, .
\end{align}
We compute $c_0$ as $\left[\overrightarrow{b_{|X|}}, \overleftarrow{b_1}\right]$. Finally, the distribution over target-side tokens at decoder step $t$ is computed as
\begin{align}
p(\hat{y}_t \given X, y_1, \ldots, y_{t-1}) = \text{softmax}(W_o c_t + b_o) \, .
\end{align}
At training time, we rely on teacher forcing to provide the token to be embedded in $e_t$. That is, we use the gold token $y_{t-1}$. At inference time, we use the generated token $\hat{y}_{t-1}$ at the previous timestep. We decode using beam search with a beam size of three, following \citet{du2017learning}.

Following \citep{zhang2016variational, jain2017creativity}, we introduce a latent variable $z \in \mathbb{R}^d$ conditional on $x$ in the decoder to model the underlying semantic space. 
We redefine the query vectors used for attention, Eq.\ (\ref{eq:query}), as
\begin{align}
k'_t &= \text{LSTM}(\left[e_{t}, c_{t-1}, z\right], k'_{t-1}) \, .
\end{align}
We introduce an approximation $q_\phi(z \given x, y)$ for the intractable true posterior $p(z \given x,y)$. Instead of the true log-likelihood, we optimize the \textit{evidence lower bound} (ELBO), a key idea underpinning variational autoencoders \cite{kingma2013auto, sohn2015learning}:
\begin{equation}
\begin{split}
\mathcal{L}(x,y; \theta,\phi) = & -\text{KL}\left(q_\phi(z \given x,y) \,\|\, p_\theta(z \given x) \right) + \\  & \mathbb{E}_{q_\phi(z \given x,y)} \left[\log p_\theta(y \given z,x)\right]
\end{split}
\end{equation}


Following \citet{zhang2016variational}, we define $q_\phi(z \given x, y)$ through a neural network with parameters $\phi$ as a Gaussian of the form
\begin{equation}
q_\phi(z \given x, y) = \mathcal{N} \left( z; \mu(x, y), \sigma(x, y)^2 \mathbf{I} \right) \, .
\end{equation}
We obtain a representation of the context paragraph used in the posterior by mean-pooling over the context encoder states. Similarly, we represent the target question using mean-pooled ELMo vectors \citep{peters2018elmo}. That is,
\begin{equation}
h_c = \frac{1}{\lvert x \rvert} \sum_{i=1}^{\lvert x \rvert} b_i ,\ \ \ \ h_q = \frac{1}{\lvert y \rvert} \sum\limits_{i=1}^{\lvert y \rvert} \text{ELMo}(y_i) \; .
\end{equation}
We then obtain the Gaussian parameters
\begin{align}
h_z = \text{ReLU}\left( W_z \; \left[h_c, h_q\right] + b_z \right) \, \\
\mu = W_\mu h_z + b_\mu,\ \ \ \log \sigma = W_\sigma h_z + b_\sigma \, .
\end{align}
Similarly, we model the conditional prior $p(z \given x)$ using a neural network
\begin{align}
p_\theta(z \given x) = \mathcal{N}(z; \mu'(x), \sigma'(x)^2 \mathbf{I}) \,\\
h'_z = \text{ReLU}(W'_z \; h_c + b'_z) \, \\
\mu' = W'_\mu h'_z + b'_\mu,\ \ \ \log \sigma' = W'_\sigma h'_z + b'_\sigma \, .
\end{align}

\begin{table*}[!htb]
\centering
\begin{tabular}{lccccc}
\toprule
system & METEOR & P-METEOR & R-METEOR & F-METEOR \\ \midrule
baseline       		& 15.98        & 15.97      &  14.02   	& 14.90             \\
 \midrule
CVAE  (25d)     		& 16.32     	& 16.29 & 14.13 & 15.02       \\ 
CVAE  (50d)     	  	& 16.31    	& 16.28 	& 14.42  & 15.11       \\
CVAE  (100d)     	 	& 16.34    	& 16.32  & 14.76  & 15.31      \\ \midrule
baseline + RBS & 16.19        & 16.19      &  14.55   	& 15.19       \\
CVAE + RBS & \textbf{16.71}    	& \textbf{16.71}  & \textbf{15.43}  & \textbf{15.93}       \\ \bottomrule
\end{tabular}
\caption{METEOR and our extensions P-, R-, and F-METEOR. METEOR scores are first averaged per context across questions and then in total across all contexts.}
\label{table:evaluation_meteor}
\end{table*}

\section{Experiments and Evaluations}\label{section:experiments}
We compare our model to the deterministic baseline from \citet{du2017learning}\footnote{The results for our implementation differ slightly from the ones in their paper. We suspect this is due to differences between our from-scratch TensorFlow implementation and their Torch-based extension of the OpenNMT framework.} on the SQuAD dataset \citep{rajpurkar2016squad}. We demonstrate comparable performance using conventional metrics, and better performance on our proposed metrics. We further corroborate this finding through human evaluation. Inspired by \citet{vijayakumar2016diverse}, we also evaluate a random beam selection (RBS) heuristic, where we induce diversity by sampling from the top $b$ beams. 
Details of hyperparameters are given in Appendix~B.



Table~\ref{table:evaluation_meteor} reports the results where METEOR is chosen as the $s$-function. METEOR has been shown to correlate well with human judgments \citep{banerjee2005meteor}. In the interest of space, results of BLEU and ROUGE are given in Appendix~C. 
For METEOR, we average first per context $c$ and subsequently over the dataset $\mathcal{D}$ to prevent overweighting contexts with more reference questions:
\begin{equation}
s(\mathcal{D}) = \frac{1}{|\mathcal{D}|}\sum\limits_{(c,Y) \in \mathcal{D}}  \frac{1}{|Y|}\sum\limits_{y \in Y} s(c,y) \, .
\end{equation}

We report results of the variational model using 25, 50, and 100 dimensional latent variables, along with the random beam selection extension of both systems the two-layer baseline and the 100-dimensional CVAE. The 100-dimensional version of our variational model performs favorably, especially in terms of F-metrics. Through our recall score, we can identify exactly where the improvement occurs --our models show larger improvement on recall, e.g. they match more references.

Increasing the dimensionality of the latent variable $z$ strictly improves the recall of our models, and consequently the F-metrics. This supports our intuition that the degree of diversity the model can express is controlled by the amount of information that is encoded in the latent variable. The RBS heuristic adds a small but rather consistent gain to both the baseline \textit{and} the CVAE model. Example outputs of our system can be found in Appendix~D.

\begin{table}[!htbp]
\centering
\begin{tabular}{l c c c}
\toprule
system      & fluency & relevancy & diversity\\ \midrule
baseline &     2.49       &      2.62            & 1.11 \\
baseline + RBS\hspace{-0.1cm} & 2.56 & 2.64 & 1.57 \\
CVAE   &     \textbf{2.57}       &     2.64        &    1.78 \\ 
CVAE + RBS & 2.47 & \textbf{2.65} & \textbf{1.86} \\ \bottomrule
\end{tabular}
\caption{Human evaluation on the test set. For each criterion, the score ranges from 1 to 3, indicating, e.g., \textit{not fluent}, \textit{almost fluent}, \textit{fluent}. Each example is rated by three annotators.}
\label{table:mturk}
\end{table}

To confirm our findings, we use Amazon Mechanical Turk to conduct human evaluation. We presented the annotators a context paragraph with questions generated by the baseline and our CVAE model. We tasked them to rate the example with three criteria, \textit{fluency}, \textit{relevancy}, and \textit{diversity}.

Table \ref{table:mturk} shows the two systems, with their RBS extension, performed comparably in terms of fluency and relevancy, whereas the CVAE demonstrated a significant improvement in diversity. (With two-sample t-tests, \textit{p}$<$0.05 for both CVAE vs.\ baseline and CVAE+RBS vs.\ baseline+RBS. More details are in the annotation file.) The finding is consistent with the automatic evaluation.






\section{Conclusion}

We have introduced a framework to extend existing evaluation metrics into $F_1$-like scoring functions explicitly rewarding diversity and enabling detailed comparison in terms of precision and recall. Furthermore, we have presented the first variational autoencoder for question generation over general text. Our model shows comparable results to the baseline in terms of conventional evaluation metrics, while producing significantly more diverse questions according to human and automatic evaluation. Our experiments suggest that the modeling of diversity is an important aspect of question generation systems, both to generate engaging questions and to better model the inherently diverse training inputs, and our development of a family of metrics suitable for evaluating models in this setting represents a step in that direction.


\bibliographystyle{acl_natbib}
\bibliography{bibliography}

\appendix

\section{In-Between Response} \label{app:in_between_response}
Our evaluation framework gives lower scores to in-between questions. We illustrate this now with an example in Section 5. 

Suppose we have a context with the reference set $R = \lbrace r_1, r_2\rbrace$, where 
\begin{itemize}
    \item $r_1$: \textit{who won \green{the 2014 world cup}}
    \item $r_2$: \textit{\pink{which event did} Germany win in 2014}
\end{itemize}

And the system outcome is a set $P$ containing a single element 
\begin{itemize}
    \item $p_1$: \textit{\pink{which event did} \green{the 2014 world cup}}
\end{itemize}

The values of METEOR, ROUGE, BLEU (sentence-level) and the corresponding F-metrics are
\begin{itemize}
    \item BLEU: 0.5946 \qquad F-BLEU: 0.2867
    \item ROUGE: 0.6240 \qquad F-ROUGE: 0.5987
    \item METEOR: 0.3773 \qquad F-METEOR: 0.3516
\end{itemize}

\section{Hyperparameters} \label{app:hyperparameters}

We implemented our models in TensorFlow. Wherever possible, we kept the hyperparameters identical to the ones used by \citet{du2017learning}. For the variational model, we experiment with different dimensionalities for the latent variable $z$, choosing from $\{25,50,100\}$.

Following \citet{sonderby2016train}, we anneal the KL term, with a scaling factor starting at $0$ and increasing at a rate of $0.03$ per iteration. This rate is determined through greedy search from the set $\{0.01, 0.03, 0.05, 0.1\}$. We apply dropouts to the latent variable $z$ on the example level, dropping out the variable entirely with a probability $0.2$, selected from the set $\{0.1, 0.2, 0.3, 0.4, 0.5\}$. For the baseline, we saw slight improvements from stacking two LSTM layers in the encoder and the decoder (with dropouts of probability $0.3$ applied in-between), while for the variational model multiple layers had no significant impact. For the RBS heuristic, we selected the number of beams $b=2$ on the development set as well.

\section{Results of Additional Metrics} \label{app:additional_metrics}
Results with ROUGE and BLEU are summarized in Table~\ref{table:evaluation_rouge} and \ref{table:evaluation_bleu}.

\begin{table*}[t]
\centering
\begin{tabular}{lccccc}
\toprule
system & ROUGE & P-ROUGE & R-ROUGE & F-ROUGE \\ \midrule
baseline       		& 37.77        & 37.61      &  33.02   	& 34.44       \\
 \midrule
CVAE (25d)     		& 37.54     & 37.19 & 33.77 & 34.91       \\ 
CVAE (50d)     	  	& 37.42   	& 37.24 & 34.01  & 35.32       \\
CVAE (100d)     	 	& 38.30    	& 38.10 & 35.07  & 36.32      \\ \midrule
baseline + RBS & 37.84        & 37.72      &  34.07   	& 35.44\\
CVAE + RBS & \textbf{38.32}    	& \textbf{38.11} & \textbf{35.24}  & \textbf{36.59}       \\ \bottomrule
\end{tabular}
\caption{ROUGE-L and our extensions P-, R-, and F-ROUGE.}
\label{table:evaluation_rouge}
\end{table*}
\begin{table*}[!htb]
\centering
\begin{tabular}{lc:ccccc}
\toprule
system & cBLEU & BLEU & P-BLEU & R-BLEU & F-BLEU \\ \midrule
baseline       						& 11.02    	& 5.17   	& 4.57    &  3.10   	& 3.55      \\\midrule
CVAE (25d)     	& 9.67 		& 4.92	    	& 4.45 	& 3.22 & 3.62       \\ 
CVAE (50d)   		& 9.78    	& 5.00   	& 4.50 	& 3.34  & 3.71       \\
CVAE (100d)  	& 10.23     & 5.15	 	& 4.67 	& 3.78  & 3.98      \\ \midrule
baseline + RBS & \textbf{11.04}    	& \textbf{5.18}   	& 4.56    &  3.12   	& 3.59\\
CVAE + RBS     & 10.39     & 5.15	 	& \textbf{4.69} 	& \textbf{3.82}  & \textbf{4.01}       \\ \bottomrule
\end{tabular}
\caption{Corpus-level BLEU-4, sentence-level BLEU-4, and our extensions P-, R-, and F-BLEU computed using the sentence-level metric. For the corpus-level metric, we sample from the CVAE once per context rather than once per reference question.}
\label{table:evaluation_bleu}
\end{table*}

\section{System Outputs} \label{app:system_outputs}

Some example outputs from the baseline system and our best-performing CVAE model are given below.

\subsection{}
{\setlength{\parindent}{0cm}
\textbf{Context:} according to the doctrine of impermanence , life embodies this flux in the aging process ,  the cycle of rebirth ( samsara ), and in any experience of loss .

\textbf{\citeauthor{du2017learning}:} what is the cycle of rebirth ?

\textbf{CVAE 1:} what is the process of impermanence ?

\textbf{CVAE 2:} according to the doctrine of impermanence , what does the cycle of rebirth mean ?

\textbf{\citeauthor{du2017learning} + RBS 1:} how does the cycle of rebirth ?

\textbf{\citeauthor{du2017learning} + RBS 2:} how does the cycle of rebirth ?

\textbf{CVAE + RBS 1:} what does the term of impermanence refer to ?

\textbf{CVAE + RBS 2:} according to the doctrine of cycle , what does the cycle of impermanence refer to ?
}












\subsection{}
{\setlength{\parindent}{0cm}
\textbf{Context:} the world trade center path station , which opened on july 19 , 1909 as the hudson terminal , was also destroyed in the attack .

\textbf{\citeauthor{du2017learning}:} on what date was the world trade center ?

\textbf{CVAE 1:} what was destroyed in the attack ?      

\textbf{CVAE 2:} when was the world trade center station opened ?

\textbf{\citeauthor{du2017learning} + RBS 1:} when did the world trade center rail open ?	

\textbf{\citeauthor{du2017learning} + RBS 2:} when was the world trade center station opened ?

\textbf{CVAE + RBS 1:} what was the hudson terminal ?	

\textbf{CVAE + RBS 2:} when did the hudson terminal open ?

}

\subsection{}
{\setlength{\parindent}{0cm}
\textbf{Context:} michiru ōshima created orchestral arrangements for the three compositions , later to be performed by an ensemble conducted by yasuzo takemoto .

\textbf{\citeauthor{du2017learning}:} who created musical arrangement for the three compositions ?

\textbf{CVAE 1:} what was the name of the orchestral ōshima created ?

\textbf{CVAE 2:} who created compositions arrangements ?

\textbf{CVAE 3:} who created created arrangements ?

\textbf{\citeauthor{du2017learning} + RBS 1:} who performed the ensemble ōshima ?

\textbf{\citeauthor{du2017learning} + RBS 2:} who performed a ensemble ōshima ?	

\textbf{\citeauthor{du2017learning} + RBS 3:} who performed the ensemble ōshima ?

\textbf{CVAE + RBS 1:} who performed the musical arrangements for the three compositions ?	

\textbf{CVAE + RBS 2:} who created the compositions to be performed by michiru ensemble ?	

\textbf{CVAE + RBS 3:} who performed musical musical in the three compositions ?
}

\subsection{}
{\setlength{\parindent}{0cm}
\textbf{Context:} examples include a concert on 23 march 1833 , in which chopin , liszt and hiller performed -lrb- on pianos -rrb- a concerto by j.s. bach for three keyboards ; and , on 3 march 1838 , a concert in which chopin , his pupil adolphe gutmann , charles-valentin alkan , and alkan 's teacher joseph zimmermann performed alkan 's arrangement , for eight hands , of two movements from beethoven 's 7th symphony .

\textbf{\citeauthor{du2017learning}:} who performed alkan 's ?

\textbf{CVAE 1:} who performed alkan 's 8th march ?	

\textbf{CVAE 2:} who performed alkan 's 8th march ?	

\textbf{CVAE 3:} how many movements did joseph play in the concert ?

\textbf{\citeauthor{du2017learning} + RBS 1:} who performed alkan 's arrangement ?

\textbf{\citeauthor{du2017learning} + RBS 2:} who performed alkan 's first solo ?

\textbf{\citeauthor{du2017learning} + RBS 3:} who performed alkan 's arrangement ?

\textbf{CVAE + RBS 1:} who performed a solo concert ?	

\textbf{CVAE + RBS 2:} who performed the concert at the march of 1876 ?	

\textbf{CVAE + RBS 3:} who performed liszt 's arrangement 23 ?

}

\end{document}